\documentclass{article}

\usepackage{microtype}
\usepackage{graphicx}
\usepackage{subfigure}
\usepackage{booktabs} 

\usepackage{hyperref}

\usepackage[accepted,noticeoff]{icml2018}

\usepackage{url}
\usepackage{amsmath,amsfonts}
\usepackage{multirow}

\usepackage{enumitem}
\setitemize{noitemsep,topsep=0pt,parsep=0pt,partopsep=0pt,leftmargin=*}
\setenumerate{noitemsep,topsep=0pt,parsep=0pt,partopsep=0pt,leftmargin=*}

\icmltitlerunning{Interpreting Deep Classifiers by Visual Distillation of Dark Knowledge}

\usepackage{xcolor}

\newcommand{\R}{\mathbb{R}}
\newcommand{\calD}{\mathcal{D}}
\newcommand{\kde}{\text{\tiny \sc KDE}}

\begin{document}

\twocolumn[

\icmltitle{
Interpreting Deep Classifiers by Visual Distillation of Dark Knowledge
}
           


\icmlsetsymbol{equal}{*}

\begin{icmlauthorlist}
\icmlauthor{Kai Xu}{ed}
\icmlauthor{Dae Hoon Park}{huawei}
\icmlauthor{Yi Chang}{huawei}
\icmlauthor{Charles Sutton}{ed,turing}
\end{icmlauthorlist}

\icmlaffiliation{ed}{School of Informatics, University of Edinburgh, Edinburgh, United Kingdom}
\icmlaffiliation{turing}{The Alan Turing Institute, London, United Kingdom}
\icmlaffiliation{huawei}{Huawei Research America, CA, USA}

\icmlcorrespondingauthor{Kai Xu}{kai.xu@ed.ac.uk}

\icmlkeywords{Machine Learning, ICML}

\vskip 0.3in
]



\printAffiliationsAndNotice{}  

\begin{abstract}
Interpreting black box classifiers, such
as deep networks, allows an analyst
to validate a classifier 
before it is deployed in a high-stakes setting.
A natural idea is to visualize
the deep network's representations, 
so as to ``see what the network sees''.
In this paper, we demonstrate that
 standard dimension
reduction methods in this setting can yield uninformative
or even misleading visualizations.
Instead, we present \textit{DarkSight}, which visually summarizes the
predictions of a classifier in a way
inspired by notion of dark knowledge. 
DarkSight embeds the data points
into a low-dimensional space such that it
is easy to compress the deep classifier
into a simpler one, 
essentially combining model
 compression and dimension reduction.
We compare DarkSight against t-SNE 
both qualitatively and quantitatively,
demonstrating that DarkSight visualizations
are more informative.
Our method additionally yields
a new confidence measure based on dark knowledge by quantifying
how unusual a given vector of predictions is.
\end{abstract}

\section{Introduction}

Despite the many well-known successes of deep 
learning \citep{bishop2006pattern,lecun1998gradient,krizhevsky2012imagenet,AMATO201347},
deep classifiers often fall short
on \emph{interpretability}, which we take to mean
whether a person can understand at a general level
why the classifier makes the decisions that it does, and
in what situations it is likely to be more or less reliable.
Lack of interpretability has been cited as a barrier
to deploying more complex classifiers, such as
random forests and deep classifiers.
The interpretability of a classifier is especially important when incorrect classifications have a high cost or when properties like fairness in a model are being verified
\citep{benitez1997artificial,caruana2015intelligible,doshi-velez17}. 

Perhaps ironically, notions of interpretability
and intelligibility are often themselves not
clearly defined \cite{lipton:mythos,doshi-velez17}. 
Our definition of interpretability is motivated by considering the
common scenario in which
a trained neural network classifier is evaluated
before deployment, which
is especially important in industry where models would influence millions of users.
Clearly, evaluating
held-out accuracy is an important first
step, but there are many more detailed questions
that we need to ask to understand
why a network makes the classifications that
it does, and in which situations the network
is most reliable, and least reliable.

A natural solution is to use visualization. For example, researchers have
proposed visualizing the activations of the hidden
layers of a deep classifier, using  
common dimension reduction techniques such as
principal components analysis \citep[PCA;][]{hotelling1933analysis}, or t-distributed stochastic neighbor embedding \citep[t-SNE;][]{maaten2008visualizing}.
This technique is described in a blog post by
\citet{karpathytsnecnn} and mentioned as a 
common
method by \citet{lipton:mythos}, although 
we are unaware of it having been studied systematically
in the research literature.
Although this is a natural idea, 
we find that it can actually provide a misleading
sense of classifier performance. 
For example, t-SNE tends 
to produce plots where points are well-separated, 
even when in fact many points lie near the decision boundary.
From such a plot, one
might infer that the classifier is confident on all points, when that is not the case. 

We propose that more reliable interpretations
of classifier performance can be made by
visualizing \emph{dark knowledge} \cite{hinton2015distilling}. 
Dark knowledge refers to the idea that the full vector
of predicted class probabilities from a deep
classifier --- not just the highest probability
output --- contains implicit knowledge that
has been learned by the classifier. 
For example, an image for which the most likely predictions, with associated 
probabilities, are \texttt{cat:0.95 dog:0.03}
is likely to be different from an image
whose predictions are \texttt{cat:0.95 car:0.03},
since dogs are more similar to cats than cars.
Dark knowledge is extracted using techniques
that have variously been called model compression or model distillation
\cite{bucila2006model,ba2014deep,hinton2015distilling}, in which a simple classifier
is trained to match the predictions of a more
complicated one. We propose that dark knowledge
can be useful for interpreting classifiers
that output probabilistic predictions.
By visualizing which data points are assigned
similar class probability vectors, an analyst
can gain an intuitive understanding of how
the classifier makes decisions.

To this end, we introduce DarkSight\footnote{Our project website \url{http://xuk.ai/darksight/} 
contains links to a PyTorch implementation of DarkSight as well as online demos for DarkSight visualizations.}, a visualisation
method for interpreting the predictions of a black-box classifier on a data set.
Essentially, DarkSight jointly performs
model compression and dimension reduction,
assigning a low-dimensional representation to each data point 
in such a way that a simpler, interpretable classifier
can easily mimic the black-box model. 
The representations and the interpretable classifier are trained end-to-end using a model compression objective, so that the full vector of predictive probabilities of the interpretable classifier on the low-dimensional data matches that of the black box classifier on the original data. 
Through a combination of detailed case studies and
quantitative evaluations, we show that DarkSight visualizations
highlight interesting aspects of the network's predictions
that are obscured by standard dimension reduction approaches.

\subsection{Design Principles}
\label{sec:design}

We identify four 
design principles, i.e. four properties, that 
low-dimensional embeddings should satisfy
to provide reliable information about a 
classifier. These not only motivate our design
but also provide a basis
for evaluating competing visualization methods
for interpretability:
\begin{enumerate}

\item \textit{Cluster Preservation.}
Points in the low-dimensional space
are clustered by the predicted class
labels, and the classifier's confidence monotonically decreases from the cluster center. 

\item \textit{Global Fidelity.}
The relative locations of clusters in the low dimensional space are meaningful. 
Clusters that are closer together correspond to classes
that are more likely to be confused by the classifier.

\item \textit{Outlier Identification}.
Data points for which the vector of predicted
class probabilities are unusual,
namely \emph{predictive outliers}, 
are easy to identify in the low dimensional space.
The classifier's predictions may be less reliable on these points
(e.g. see Figure~\ref{fig:res-mnist}).

\item \textit{Local Fidelity.}
Points that are nearby in the low dimensional space have 
similar 
predictive distributions according to the classifier.
\end{enumerate}

Most nonlinear dimension reduction techniques like t-SNE
satisfy local fidelity by design, but in \autoref{sec:eval} we show that they often
fall short on the other principles, and can
mislead the analyst as a result.

\section{Related Work}
 
DarkSight combines ideas from knowledge
distillation, dimension reduction, and visualization and interpretation
of deep networks. We 
review work in each of these areas.

\textbf{Knowledge distillation}. 
Knowledge distillation \cite{bucila2006model,DBLP:journals/corr/BastaniKB17a,ba2014deep,hinton2015distilling} means training one model, called
a \emph{student model}, to generalize
in the same way as a second \emph{teacher model}, which is usually more complex. This is
also called \emph{model compression}, because
the teacher model is compressed into the student.
Interestingly, the relative probabilities predicted by the teacher
for lower-ranked classes contain important information
about how the teacher generalizes
\cite{bucila2006model}. For example, consider two handwritten digits that are predicted as 7's--- knowing whether the second-best prediction is 2
or 1 is highly informative.
The implicit knowledge that is represented by the full vector
of predicted
class probabilities	
has sometimes been referred to as \emph{dark knowledge}
learned by the network \citep{hinton2015distilling}. 
Model compression was originally proposed to reduce
the computational cost of a model at runtime
\cite{bucila2006model,ba2014deep,romero2014fitnets},
but has later been applied for interpretability
(see below).

\textbf{Dimension reduction.} Visualization of high-dimensional data, such as  
predictive
probabilities, necessarily involves \emph{dimension reduction}
of the data to a low-dimensional space.
This has been an important topic in visual analytics over the last few decades \citep{de2003visual}.
Classical methods from statistics include principal 
components analysis 
\citep{hotelling1933analysis} and 
multidimensional scaling \cite{cox2000multidimensional}.
More recently, t-SNE has been exceptionally
popular for text and image data
\citep{maaten2008visualizing}.
For a review of more recent approaches to dimension reduction,
see \citet{van2009dimensionality}.
\textbf{Interpreting deep networks.} 
Various methods have been proposed in recent years to interpret 
neural networks \citep{lipton:mythos}.
These include 
compressing neural networks into simple models, e.g. a small decision tree, which is easy to interpret \citep{craven1996extracting},
retrieving typical inputs so that one can interpret by examples via hidden activations \citep{caruana1999case} or influence functions \citep{2017arXiv170304730K}
and generating artifacts or prototypes to explain model behaviour \citep{Ribeiro2016lime,2015arXiv150301161K}.
In contrast to deep neural networks, 
an alternative is to restrict the models to belong to a family that is
inherently easy to interpret \citep{caruana2015intelligible,2014arXiv1410.4510D,letham2015interpretable}.



\textbf{Visualizing deep networks.}
Another way to interpret deep neural networks
is by means of visualization; for an overview,
see \citet{olah2017feature}.
First, \emph{feature visualization} methods visualize different layers learnt by neural networks,
for example, by producing images that
most activate individual units,
ranging from low-level features like edges and textures to 
middle-level features like patterns and 
even high-level concepts like objects
\citep{erhan2009visfeatures,mahendran2015understanding,oalh2015feature,nguyen2015deep}.
An alternative is
 \emph{attribution methods} that visualize how different parts of the input contribute to the final output,
 such by generating sensitivity maps over the input \cite{10.1371/journal.pone.0130140,baehrens2010explain,selvaraju2016grad,smilkov2017smoothgrad}.
Although these methods can produce
informative visualizations, it is difficult to
scale these displays to large networks and large validation sets
--- where by ``scalability'' we are referring
not to the computational cost of generating the display, but whether the display
does not become so large that a person cannot
examine it.

Most closely related work to ours is the proposal by
\citet{karpathytsnecnn} to
apply t-SNE to the features from the second to last layer in a deep classifier,
producing a two-dimensional embedding in which nearby
 data items have similar high-level features
 according to the network.
We will observe that these plots can be misleading because they contain well-separated clusters 
even when, in fact, there are many points nearby the decision boundary (see Section~\ref{sec:eval}).


\section{DarkSight}
\label{sec:darksight}
The goal of DarkSight is to interpret the predictions of a black-box classifier
by visualizing them in a lower dimensional space.
Our method takes as input an already-trained classifier, such as a deep network,
to be interpreted; we call this the \emph{teacher classifier}. Our method
relies on the teacher producing a probability distribution $P_T(c | x)$ over classes
rather than a single prediction. We are also given a validation
set of data points $\calD_V = \{(x_i, c_i)\}$ separate from the data
used to train the model, and the task is to 
visually summarize the predictions made
by the teacher on $\calD_V$.

Our method combines dimension reduction
and model compression.
For each data point $x_i,$ we define
the \emph{prediction vector} $\pi_i = P_T(c_i | x_i)$
produced by the teacher. Our goal will be to represent each point $x_i$ 
in the visualization by a low-dimensional embedding $y_i$.
To do this, we train an interpretable \emph{student classifier}
$P_S(\cdot| y; \theta)$ in the lower dimensional space,
where $\theta$ are the classifier parameters.
The training objective is a model compression 
objective to encourage the student prediction vector
$P_S(c_i | y_i; \theta)$
to match the teacher's prediction vector
$\pi_i = P_T(c_i | x_i)$.
Importantly, we optimize the objective jointly
with respect to both the classifier parameters $\theta$
and also the embeddings $Y = \{y_i\}$.

This perspective highlights the key technical novelties
of our approach. Compared to previous work
in model compression, we optimize both the
parameters \emph{and the inputs} of the student
model. Compared to previous work on dimension reduction, the embeddings
$y_i$ can actually be viewed as representations of the prediction vectors
$\pi_i$ rather than the original input $x_i$.
To justify this interpretation, observe
that the optimal choice of $y_i$ according
to the objective function \eqref{eq:obj} below depends on $x_i$ only via $\pi_i$. 



\subsection{Objective}

To formalize the idea of ``matching dark knowledge''
between the student and the teacher, we
minimize an objective function that encourage
matching the 
 the predictive distribution 
 of the teacher with that of the student, namely,
\begin{equation}
    L(Y, \theta) = \frac{1}{N} \sum_{i=1}^N D(P_T(\cdot | x_i), P_S(\cdot | y_i; \theta)),
    \label{eq:obj}
\end{equation}
where $D$ is a divergence between probability
distributions.

Instead of using common choices for $D$ suggested in the model compression literature, such as 
KL divergence and Jensen-Shannon (JS) divergence \citep{papamakarios2015},
we empirically found that more informative visualizations
arose from the symmetric KL divergence
\begin{equation}
    KL_{sym}(P, Q) = \frac{1}{2} (KL(P, Q) + KL(Q, P)),
    \label{eq:sym_kl}
\end{equation}
where $KL(P, Q) = -\sum_{k=1}^K P(k) \log \frac{Q(k)}{P(k)}$.

\subsection{Choice of Student Model}
\label{sec:student}


Interestingly, we achieve good results while
using a fairly simple choice of student
model $P_S(c_i = k | y_i; \theta)$.
We suggest that because we optimize with respect to both the student classifier's parameters and inputs,
i.e. the embeddings,
even a simple student classifier
has substantial flexibility to mimic the
teacher.
We use the naive Bayes classifier
\begin{equation}
    P_S(c_i = k | y_i; \theta) = \frac{P(y_i | c_i = k; \theta_c) P(c_i = k; \theta_p)}{P(y_i|\theta)}.
    \label{eq:cond}
\end{equation}
Naive Bayes  has several advantages in our setting.
First, since it models data from each class independently, 
it encourages the low-dimensional embeddings to be separate clusters.
Additionally, we can choose the class-conditional distributions
to be more easily interpretable.


Two natural choices for the distribution $P(y_i | c_i = k; \theta_{c})$ are the Gaussian and 
Student's $t$-distribution.
Compared to the Gaussian, the $t$-distribution encourages 
the low-dimensional points to be more centred
because of its heavy tail property.
This is related to the ``crowding problem'', e.g. a sphere with radius $r$ in high dimension is hard to map to a sphere with the same radius $r$ in low dimension \citep{maaten2008visualizing}.
Also, visually we want to encourage
the mean parameters $\mu$ for each class 
to lie within the corresponding low-dimensional points for that class.
However, we empirically found that this does not
always happen with  the Gaussian.
Therefore we choose
    $P(y_i | c_i = k; \theta_{\text{c}}) = t_\nu(y_i; \mu_k, \Sigma_k)$,
where $t_\nu(y_i; \mu_k, \Sigma_k)$ is a non-centered Student's 
t distribution with mean $\mu_k$, covariance matrix $\Sigma_k$, and $\nu$ degrees of freedom.
The prior over the classes is modelled by a categorical distribution
    $P(c_i = k; \theta_{p}) = \mathcal{C}\text{at}(c_i = k; \sigma(\theta_p)),$
where $\sigma$ is the softmax and $\theta_p \in \R^K$ are parameters.

\subsection{Training}


The training of DarkSight is done with stochastic gradient descent (SGD) on \eqref{eq:obj}.
We experiment both \textit{plain SGD} on $\{Y, \theta\}$ and 
\textit{coordinate descent by SGD} for $Y$ and $\theta$, 
the latter of which gives slightly better results with the sacrifice of
doubling running time.
We also find that using techniques like annealing can help avoid poor local optima.

For special cases, there can exist more efficient learning algorithms, 
for example if logistic regression is
used as the student and MSE error is used
as the loss;
see Appendix~\ref{app:svd}.

\subsection{Confidence Measure}
\label{sec:confidence}

A byproduct of DarkSight is that the low-dimensional representation
allows us to define new confidence measures for the teacher.
Intuitively, we should have less confidence
in the teacher's prediction if the full prediction vector is unusual
compared to the other points in the validation set.
To formalize this intuition, we can
estimate
the density of the embeddings $y_i$, using a standard method like kernel density estimation.
This yields an estimate
$\hat{p}_\kde(y_i)$ that we can use as
a measure of confidence in the teacher's
prediction of $x_i$.
Although one might consider
performing density estimation
directly in the space
of prediction vectors $\pi_i$,
this is a much more difficult problem,
because it is a density estimation problem with a simplex constraint, 
where common methods may over-estimate near the simplex boundary.
However, as the embeddings $y_i$ 
are optimized to retain information about 
$\pi_i$, it is reasonable 
and simpler to use $\hat{p}_\kde(y_i)$
as a measure of whether $\pi_i$ is unusual.

We contrast our confidence measure with the commonly
used \textit{predictive entropy}
$H[P_T(\cdot | x_i)]$.
The predictive entropy
does \emph{not} take dark knowledge
into account because it 
 does not
consider correlations between class predictions. 
For example, consider two prediction vectors
$\pi_1 = \texttt{[cat:0.95 dog:0.03 ...]}$ and 
$\pi_2 = \texttt{[cat:0.95 airplane:0.03 ...]}$. 
Intuitively, we should
have lower confidence in $\pi_2$ 
because confusing cats with dogs makes more sense than confusing cats with airplanes.
Despite this intuition, $\pi_1$ and $\pi_2$ have the same
predictive entropy, because the entropy
is invariant to relabeling. Confidence measures based on dark knowledge, however,
can treat these two prediction vectors differently.



\section{Evaluation}
\label{sec:eval}

We use DarkSight to visualize classifiers on three datasets: 
a LeNet \citep{lecun1998gradient} with 98.23\% test accuracy on MNIST \citep{lecun1998mnist}, 
a VGG16 \citep{simonyan2014very} with 94.01\% test accuracy on Cifar10 \citep{krizhevsky2014cifar} and 
a wide residual network (Wide-ResNet) \citep{zagoruyko2016wide} with 79.23\% test accuracy on Cifar100 \citep{krizhevsky2009learning}.
For each case we visualize a separate set of 10,000 instances, 
which is not used to train the classifiers.
Full details of intializations and 
hyperparameters 
are given in Appendix~\ref{app:exp-setup}.

We compare the DarkSight
visualization to standard 
dimension reduction approaches.
Typically, it is 
 considered desirable to produce
a visualization with clearly defined clusters,
because this means the method has identified structure
in the data. But as we will see, such a visualization
can be misleading for interpretability if important information
about the prediction vectors is missing.
Instead, we will evaluate different visualization methods
along
the four design properties described in Section~\ref{sec:design}:
local fidelity, cluster preservation, global fidelity and outlier identification.


We compare against the t-SNE method \cite{maaten2008visualizing} because
it is widely adopted for dimension reduction,
and has previously been proposed for interpreting
the deep classifiers \cite{karpathytsnecnn}.
In preliminary experiments, we attempted to apply PCA as well,
but the results were disappointing, so we omit them for space.
We consider several different methods for interpretability, each of which applies t-SNE
to different inputs:
\begin{enumerate}
	\item  \textit{t-SNE prob} uses the predictive probability vectors;
    \item \textit{t-SNE logit} uses logits of the predictive probability vectors, i.e. the last layer just
   before the softmax;
    \item \textit{t-SNE fc2} uses the final feature representations of the inputs, i.e. the layer before logit. (In LeNet, this is the second fully connected layer, called fc2.)
\end{enumerate}
All t-SNEs above are trained for 1,000 epochs with a batch size of 1,000 on the same dataset as DarkSight. 
We try perplexities from 2 to 80 when training t-SNE and 
we only report the best one from each t-SNE for each evaluation.

\subsection{Cluster Preservation}

DarkSight has a good in-cluster property: points close to the cluster center have higher confidence than points away from the cluster center. 
This is important when using the visualization to check for outliers, 
i.e. points which the classifier is uncertain of.
To see this, we
use the predictive entropy $H[\pi_i]$
as a confidence measure,
a common choice in information theory
\citep{Gal2016Uncertainty}, to color points.
\begin{figure}[ht]
    \centering
    \subfigure[DarkSight]{
        \includegraphics[clip, trim=1.0cm 1.0cm 1.0cm 1.0cm, width=.29\columnwidth]{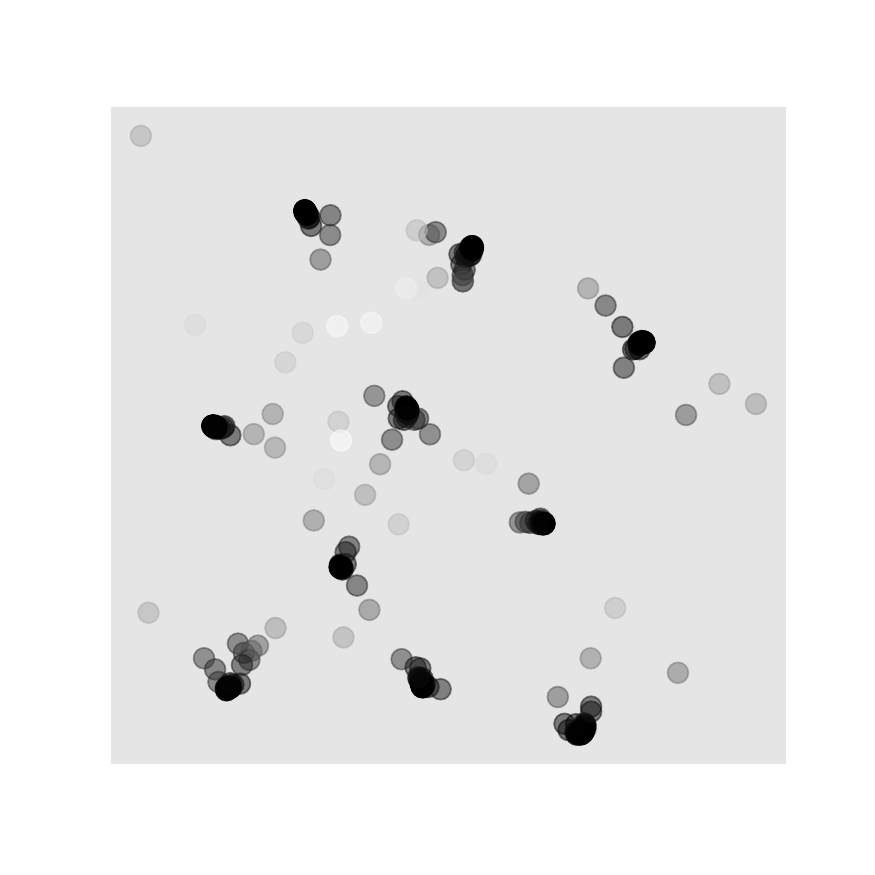}
        \label{fig:eval-cluster-ds}
    } 
    ~
    \subfigure[t-SNE prob]{
        \includegraphics[clip, trim=1.0cm 1.0cm 1.0cm 1.0cm, width=.29\columnwidth]{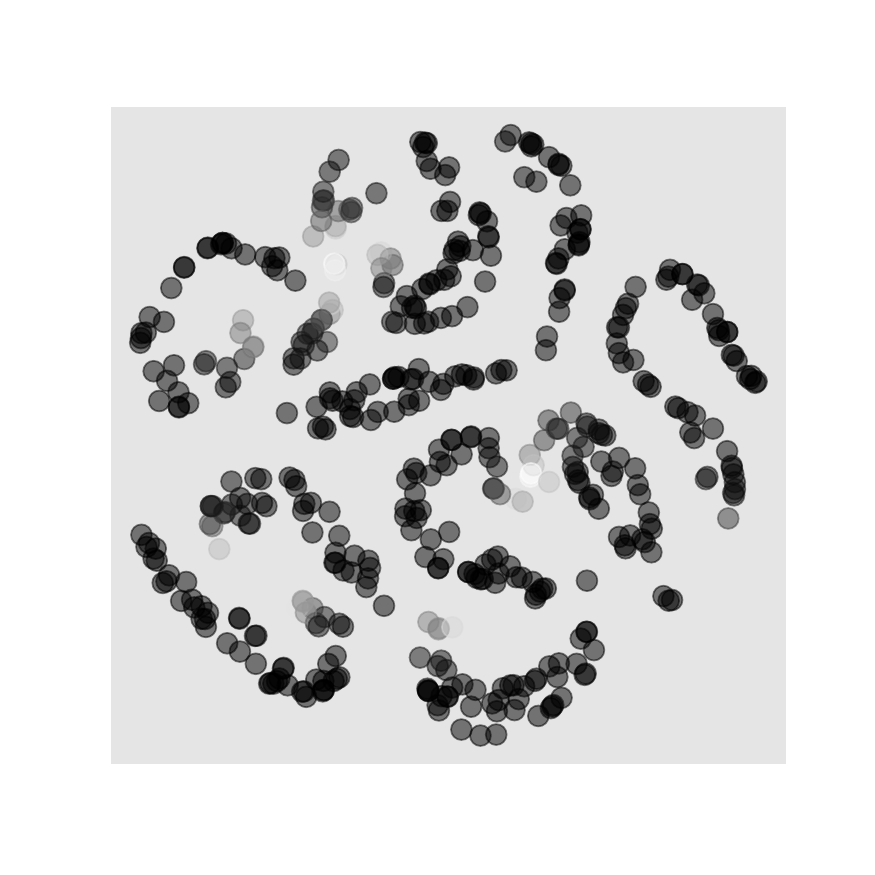}
        \label{fig:eval-cluster-p}
    }
    ~
    \subfigure[t-SNE logit]{
        \includegraphics[clip, trim=1.0cm 1.0cm 1.0cm 1.0cm, width=.29\columnwidth]{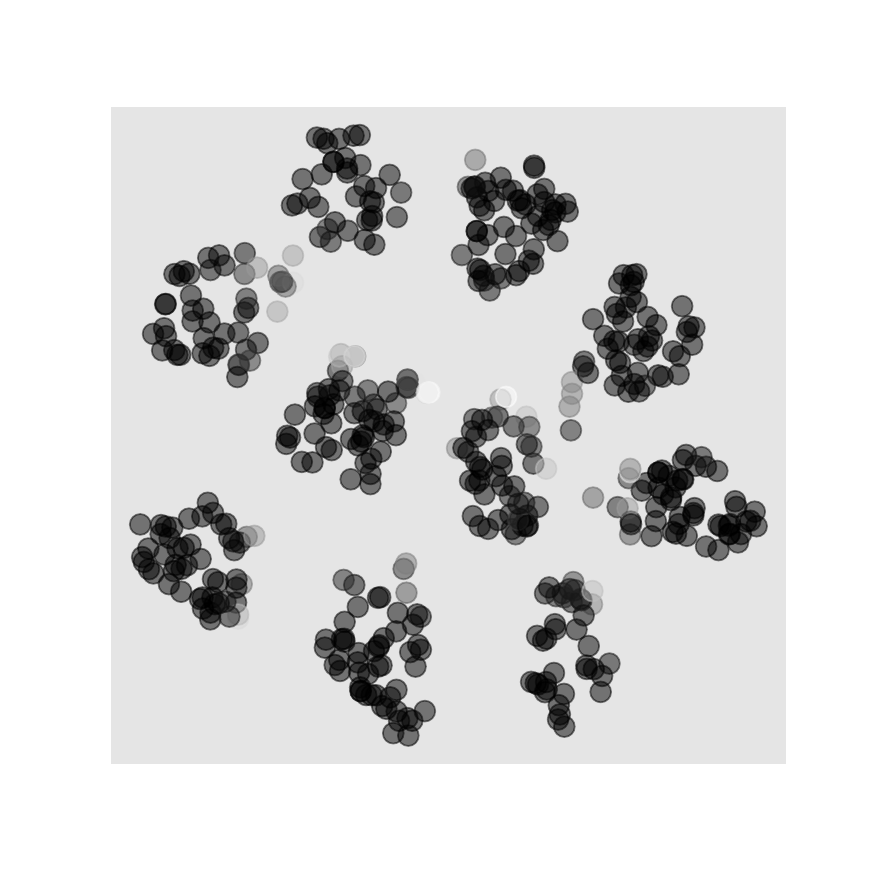}
        \label{fig:eval-cluster-logit}
    }
    \caption{Scatter plots with points colored by predictive entropy. Dark points have large values. In all three plots, the same random subset (500 out of 10000) points is shown.}
    \label{fig:eval-cluster}
\end{figure}
Figure~\ref{fig:eval-cluster} shows the scatter plots from LeNet on MNIST with each point shaded by its confidence.
It can be seen that DarkSight puts points with high predictive confidence in the cluster centers while t-SNE tends to spread points with high and low confidence throughout the cluster.
An analyst might naturally interpret
points near the centre of a cluster as more
typical, and might look to the edges of a
cluster to find unusual data points that the classifier fails on. 
This interpretation would fail for the
t-SNE visualizations,
so we would argue that they are, in this aspect, misleading.

Note that a direct result of this property is that 
points in midway of two clusters are similar to both corresponding classes
(see digits labelled as Case 2 in Figure~\ref{fig:res-mnist}).
\begin{figure}[ht]
    \centering
    \subfigure[t-SNE prob]{
        \includegraphics[clip, trim=1.0cm 1.0cm 1.0cm 1.0cm, width=.45\columnwidth]{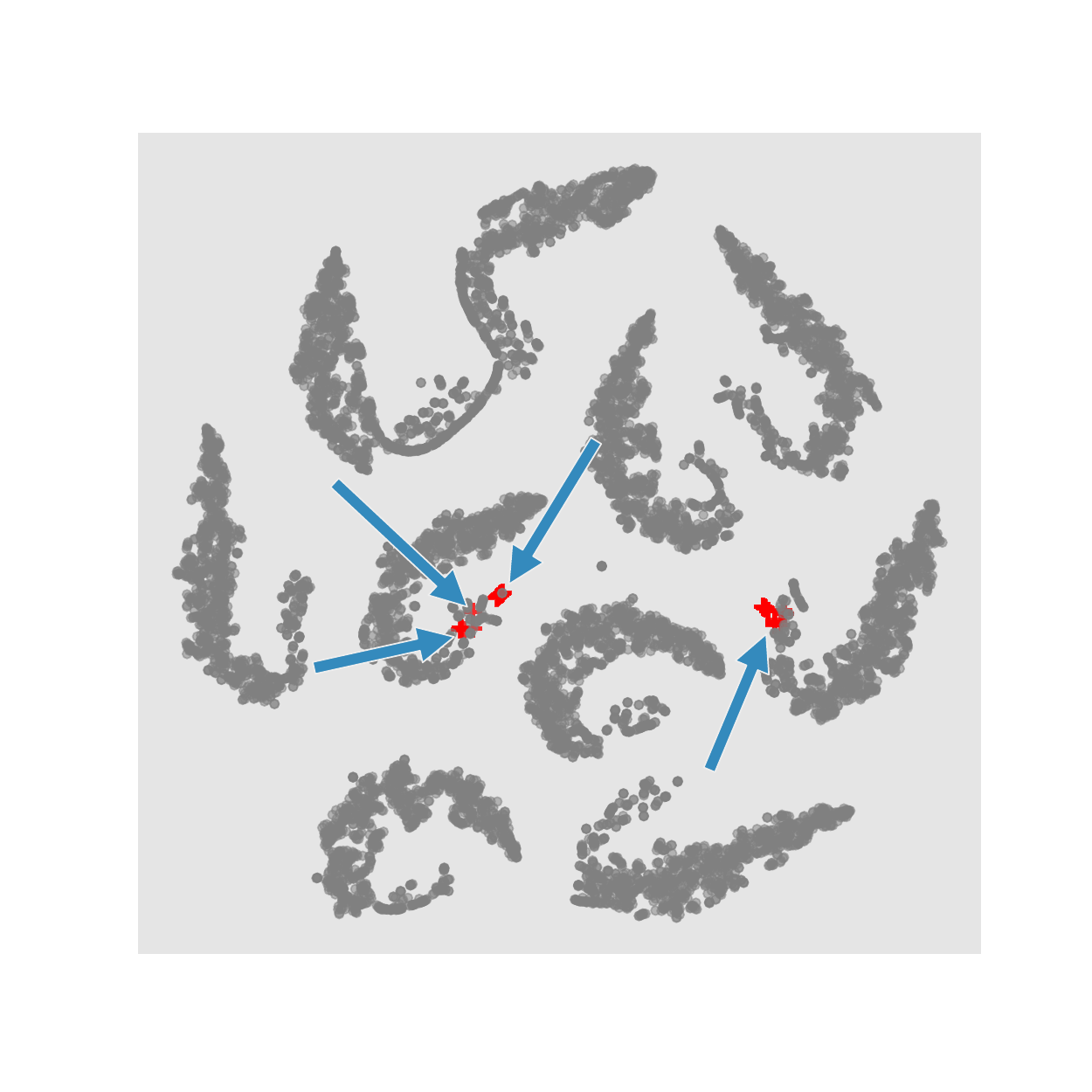}
        \label{fig:eval-cluster-p-cifar10}
    }
    ~
    \subfigure[t-SNE logit]{
        \includegraphics[clip, trim=1.0cm 1.0cm 1.0cm 1.0cm, width=.45\columnwidth]{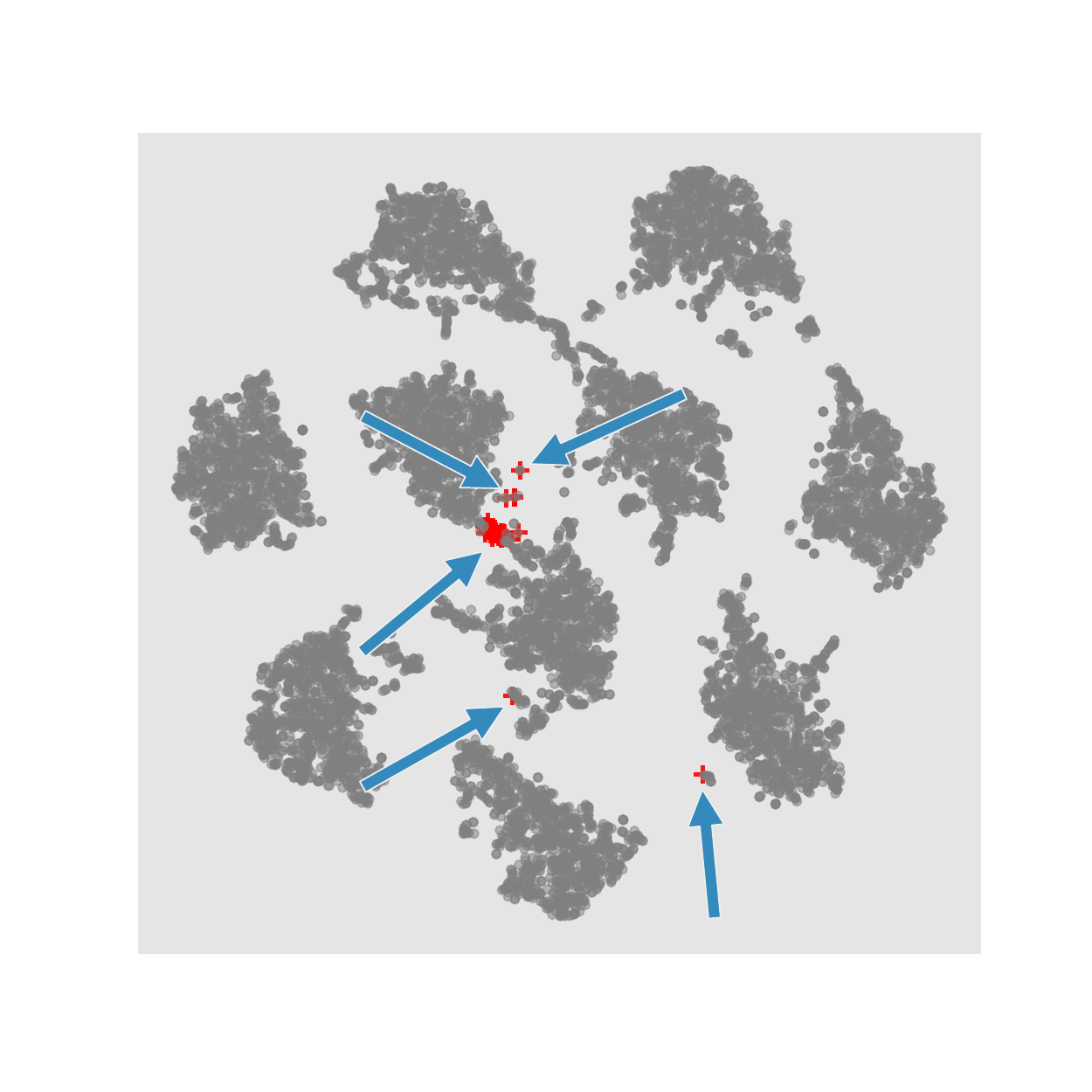}
        \label{fig:eval-cluster-logit-cifar10}
    }
    ~
    \subfigure[Predictive probabilities of points in the black box]{
            \includegraphics[clip, trim=1.4cm 0.4cm 1.4cm 0.4cm, width=.99\columnwidth]{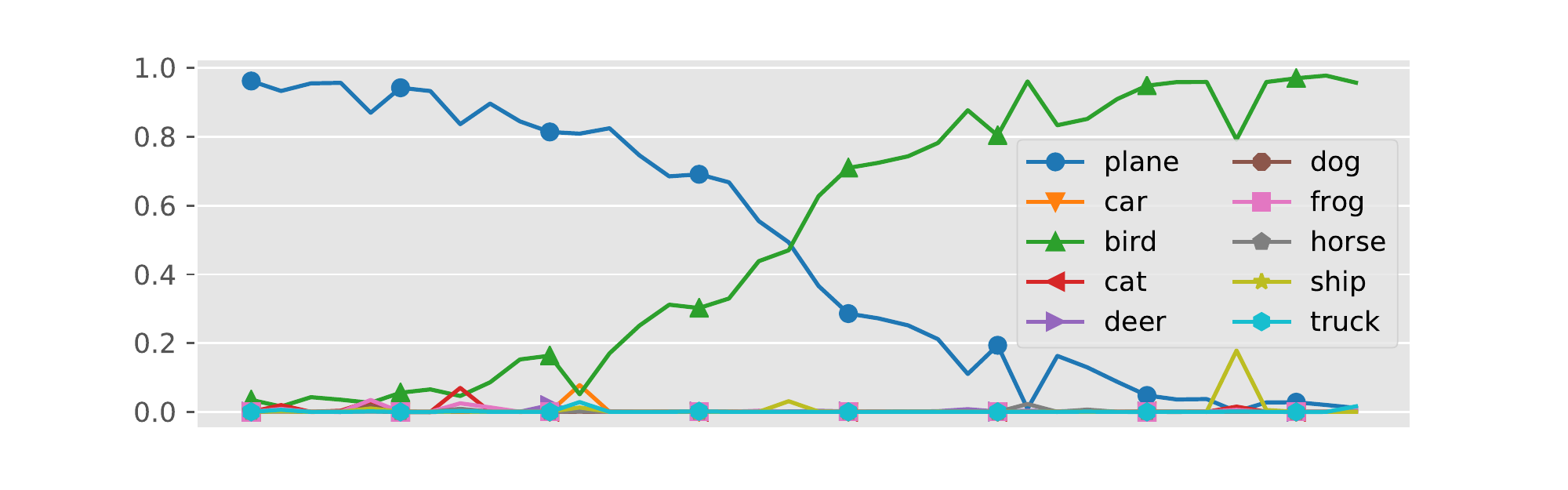} \label{fig:black-box}
        \label{fig:eval-cluster-trace-cifar10}
    } 
    \caption{Correspondence points in the black box of Figure~\ref{fig:res-cifar10}. 
    (a) and (b) are t-SNE plots with these points colored by red; red areas are also pointed by blue arrows. 
    (c) is the plot of predictive probabilities of each class for these points; left to right in x-axis corresponds to points in the box from left to right along the curve.
    }
    \label{fig:eval-cluster-box}
\end{figure}
Also refer to Figure~\ref{fig:res-cifar10} where several pairs clusters are directly adjacent
along a curve.
This happens because the predictive vectors along this curve have two top predictive probabilities that dominating the others,
and the values of the top two probabilities smoothly interchange with each other along the curve.
For example, consider the points in the in the black box of Figure~\ref{fig:res-cifar10}.
The DarkSight visualization suggests that points
along this curve smoothly transition from predictions
of planes to predictions of birds. In
Figure~\ref{fig:eval-cluster-trace-cifar10}, we zoom in to the points within
the box, and we find that this smooth transition 
is exactly what is happening. So the DarkSight visualization correctly reflected the model's predictions.
However, as shown by corresponding points in Figure~\ref{fig:eval-cluster-p-cifar10} and Figure~\ref{fig:eval-cluster-logit-cifar10},
such transition between clusters are 
hardly visible
 in t-SNE plots.
There are 38 such instances in the box which are uncertain between bird and plane,
which t-SNE over-plots uncertain points in a small area. This is unfortunate
for interpretability, as such points are precisely those which are most of interest
to the analyst.


\subsection{Global Fidelity}


In both DarkSight and t-SNE plots, nearby clusters are more likely to be "predicted together", 
i.e. if the classifier is likely to confuse between two classes,
both methods tend to position them as nearby clusters.\footnote{
    One can verify this by examining the confusion matrix.
    The confusion matrix for Cifar10 is given in Table~\ref{tab:conf-mat-cifar10} (Appendix~\ref{sec:conf-mat-cifar10}).}
But this does not always hold.
Sometimes, for both DarkSight and t-SNE,
commonly-confused classes clusters are not placed next to each other visually, or neighboring
clusters are not commonly confusable.
However, for DarkSight we can gain more insight
by examining the difference between the clusters.
We have observed then when two neighboring clusters
are directly adjacent, like the bird and plane
clusters in Figure~\ref{fig:res-cifar10}, 
then they tend always to be confusable clusters. We do not observe this useful
phenomenon in the t-SNE plots.

Moreover, DarkSight visually shows global patterns based on
the arrangement of the clusters, which is not the case for t-SNE.
For instance in Figure~\ref{fig:res-cifar10}, it can be seen
that the lower right clusters (pink, red, gray, brown, purple and green) form a group 
while the upper left clusters (ocean, orange, yellow and blue) form another.
In fact, the lower right classes are all animals and upper left are all vehicles.\footnote{
    We generate multiple DarkSight plots and
    this finding is almost consistent on different runs, with only small differences.}
More interestingly, the only two classes from each group that have a curve between them are ``bird'' and ``plane'',
which are semantically similar to each other.


\subsection{Outlier Identification}
\label{sec:eval-uncertainty}

As discussed in Section~\ref{sec:confidence}, 
the density of the DarkSight embeddings can be useful for detecting outliers
because of the idea of dark knowledge.
In this section, we evaluate quantitatively
whether DarkSight outliers tend to correspond
to less reliable predictions; later, in Section~\ref{sec:interpret}, we will evaluate
this qualitatively.
To obtain density estimations on embeddings,
we experiment with two methods: kernel density estimation (KDE) and Gaussian mixture estimation (GME);
we only report the one that gives better performance in each evaluation.
We evaluate the effectiveness of 
a confidence measure by measuring if the classifier
is more accurate on higher-confidence predictions. In particular, when the confidence is below $\delta$, we allow the classifier to \emph{reject
a point}, i.e.
decline to make a prediction without paying
a penalty in accuracy. We compare
confidence measures by
an accuracy-data plot, which 
the accuracy of the thresholded classifier when
forced to predict for a given
percentage of the data.




First, we run density estimation on the low-dimensional embeddings produced
by both t-SNE and DarkSight,
to show that the density of the DarkSight embeddings is a more useful confidence measure.
Second, to get a sense of how
much information about the global density is lost in a 2D embedding, we
run density estimation in the original space of prediction vectors, $P_T(\cdot|x_i)$, to attempt to estimate an upper bound
on the performance of methods based on a 2D visualization.
Also, because the prediction vectors are in a high-dimensional simplex, 
we can model the vectors using a mixture of Dirichlet, 
which gives density that we call Dirichlet mixture estimation (DME).

\begin{figure}[ht]
    \centering
    \subfigure[MNIST]{
        \includegraphics[width=.463\columnwidth]{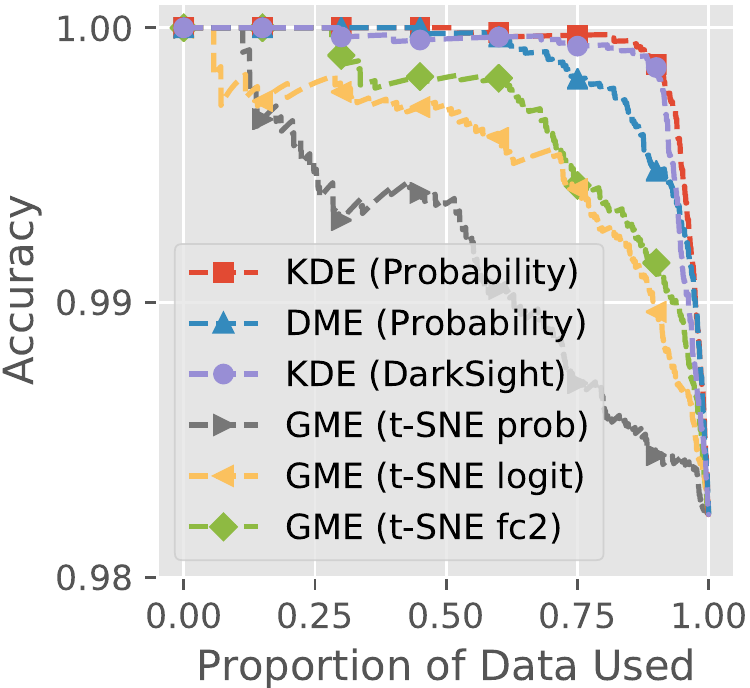}
    }
    ~
    \subfigure[Cifar10]{
        \includegraphics[width=.463\columnwidth]{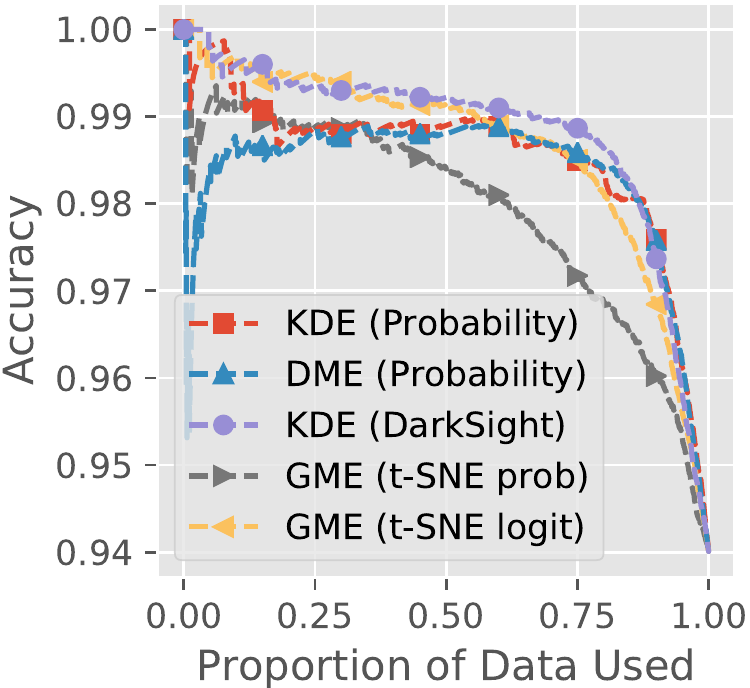}
    }
    \caption{Data-accuracy plot for different confidence measures.
    }
    \label{fig:precision-recall}
\end{figure}

The accuracy-data plots in Figure~\ref{fig:precision-recall}, for MNIST (LeNet) and Cifar10 (VGG16),
compare the effectiveness of different confidence measures 
by showing how the performance changes when different amount of points are rejected by different thresholds.
Ideally we would hope only those data points on which the model may make mistakes 
are rejected, i.e. the curves in Figure~\ref{fig:precision-recall} should be close to the upper right corner of the plot.
Overall, in both figures, KDE on DarkSight embeddings gives good results consistently,
generally outperforming the other methods based on low-dimensional embeddings.
This supports our statement that DarkSight preserves dark knowledge in the low-dimensional space 
and the density estimation on it is an effective confidence measure.
For both figures, in regions where the most of data is used, the results from KDE (DarkSight) and KDE (probability)
are similarly best.
It is also interesting to note that, for Cifar10, KDE (DarkSight) outperforms two density estimations on the original
probability space, which indicates that DarkSight captures some information about confidence
that direct density estimations on probability fail to capture.

In future work, this outlier detection method could enable a tool where the analyst can visually
mark portions where the classifier is unreliable and have the classifier
refuse to make a decision in those areas of the space.
A further interesting work would be how to interactively learn the confidence measure,
based on a small amount of user feedback.



\subsection{Local Fidelity}
\label{sec:eval-local}
In order to quantitatively evaluate how well the predictive probability is preserved locally in 2D, 
we define a metric based on k-nearest neighbours (kNNs) as below
\begin{equation}
    M_k(Y) = \frac{1}{N} \sum_{i=1}^{N} \frac{1}{k} \sum_{j \in \text{NN}_k(y_i)} JSD(p_i, p_j),
    \label{eq:knn-metric}
\end{equation}
where $p_i = P_S(\cdot | y_i)$, $JSD$ is the Jensen-Shannon distance (JSD)
\footnote{JSD is defined as the square root of JS divergence $JSD(P,Q) = \sqrt{JS(P,Q)}$, where
$JS(P,Q) = \frac{1}{2}(KL(P,M) + KL(Q,M))$ and $M = \frac{1}{2}(P + Q)$. JSD is used here because
it is a metric and its underlying JS divergence is neither the divergence used by DarkSight nor t-SNE, i.e. it is fair.} and
$\text{NN}_k(y_i)$ is the set of the indices of the $k$-nearest neighbours of $y_i$ in the 2D embedding.

\begin{figure}[ht]
    \vskip 0.2in
    \begin{center}
        \includegraphics[width=0.95\columnwidth]{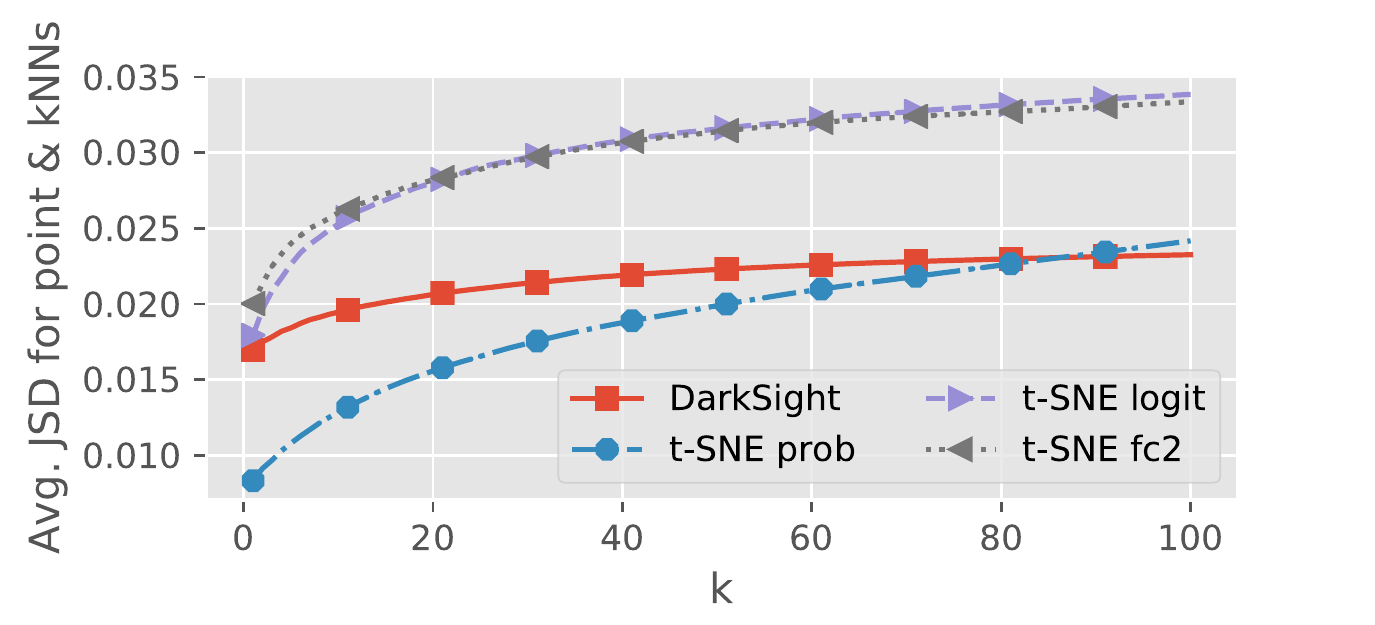}
        \end{center}
    \vskip -0.2in
    \caption{Local fidelity $M_k(Y)$ on MNIST as a function of the number of neighbours $k$,
     for DarkSight and t-SNE visualizations.
     Note: t-SNE prob is being optimized specifically for local fidelity.
    } 
    \label{fig:eval-local-knn}
\end{figure}

Figure~\ref{fig:eval-local-knn} shows the local fidelity
performance 
of LeNet on MNIST
with the number of neighbours $k$ varying,
where smaller values are better.
It can be seen that both DarkSight and t-SNE prob 
do a better job on preserving predictive probability locally
than t-SNE logit and t-SNE fc2.
In other words, the performance of t-SNE depends very 
much on which quantities are visualized.
This is because early layers of LeNet do not have much
discriminative information but only hierarchical features.
Also notice that
the fidelity $M_k(Y)$ of t-SNE is better for low $k$ 
while that of DarkSight is better for high $k$. 
This is 
because
t-SNE is primarily a method for producing embeddings with local fidelity 
while the DarkSight objective function is more of a global measure.
As we have shown in previous subsections, 
optimizing
this global objective provides benefits that t-SNE lacks.


\subsection{Computational Efficiency}
The time complexity of DarkSight is $O(N)$, 
assuming the underlying automatic differentiation implementation has constant time complexity.
On the other hand, t-SNE
has a computational complexity of $O(N^2)$,
as it requires the calculation of pairwise distances.
DarkSight also scales better than the
 faster $O(N \log N)$ variants of t-SNE \citep{van2014accelerating,linderman2017efficient}.
For empirical results on running time, see Appendix~\ref{app:time}.

\subsection{Summary of Properties}

Table~\ref{tab:summary} gives a summary of the methods evaluated above, regarding the four properties and time complexity.
\begin{table}[t]
    \caption{Comparisons between DrakSight and t-SNE. 
    Properties are defined in Section~\ref{sec:design}.
    $\surd$: good, $\times$: bad and $\sim$: acceptable.
    }
    \vskip 0.15in
    \begin{center}
        \begin{small}
            \begin{sc}
                \begin{tabular}{lccccl}
                    \toprule
                    Method / Property   & 1         & 2         & 3         & 4         & Time      \\
                    \midrule 
                    DarkSight           & $\surd$   & $\surd$   & $\surd$   & $\sim$    & $O(N)$    \\ \cline{6-6} 
                    t-SNE prob          & $\sim$    & $\times$  & $\times$  & $\surd$   & \multirow{3}{1.4cm}{$O(N^2)$ or $O(N\log N)$}\\
                    t-SNE logit         & $\times$  & $\sim$    & $\sim$    & $\times$  & \\
                    t-SNE fc2           & $\times$  & $\times$  & $\times$  & $\times$  & \\
                    \bottomrule
                \end{tabular}
            \end{sc}
        \end{small}
    \end{center}
    \vskip -0.1in
    \label{tab:summary}
\end{table}
Among all methods, DarkSight is the only one that has all properties of interest and scales well.

\subsection{Quality of Model Compression}

\begin{table}[t!]
    \caption{Training results of DarkSight for different datasets. 
    Note: Acc\#ground is the accuracy towards true labels and Acc\#teacher is the accuracy towards
    the predictions from the teacher.}
    \vskip 0.15in
    \begin{center}
        \begin{small}
            \begin{sc}
                \begin{tabular}{llcc}
                    \toprule
                    Dataset     & $KL_{sym}$    & Acc\#ground   & Acc\#teacher  \\
                    \midrule 
                    MNIST       & 0.0703        & 98.2\%        & 99.9\%        \\
                    Cifar10     & 0.0246        & 94.0\%        & 99.7\%        \\
                    Cifar100    & 0.383         & 79.2\%        & 99.9\%        \\
                    \bottomrule
                \end{tabular}
            \end{sc}
        \end{small}
    \end{center}
    \vskip -0.1in
    \label{tab:results}
\end{table}

Finally, it is reasonable to wonder if it is
even possible to obtain high-fidelity student
models when the embedding space is restricted to
two dimensions.
Table~\ref{tab:results} shows the quality of model compression that was achieved, in
terms of both symmetric KL divergence and accuracy
of the student at matching the teacher's predictions.
For the tasks with 10 classes (MNIST and Cifar10), the optimization objective
$KL_{sym}$ reaches very small values, and for the Cifar100 task, the value is still reasonable.
We conclude that the student models
are successful at matching the teacher's predictions.



\section{Case Studies}
\label{sec:interpret}

\begin{figure*}[!ht]
    \centering
    \begin{tabular}[c]{lc}
        \multirow{2}{*}[2.07in]{\subfigure[LeNet on MNIST]{
                \includegraphics[clip, trim=4.7cm 4.7cm 3.7cm 4.7cm, width=.69\textwidth]{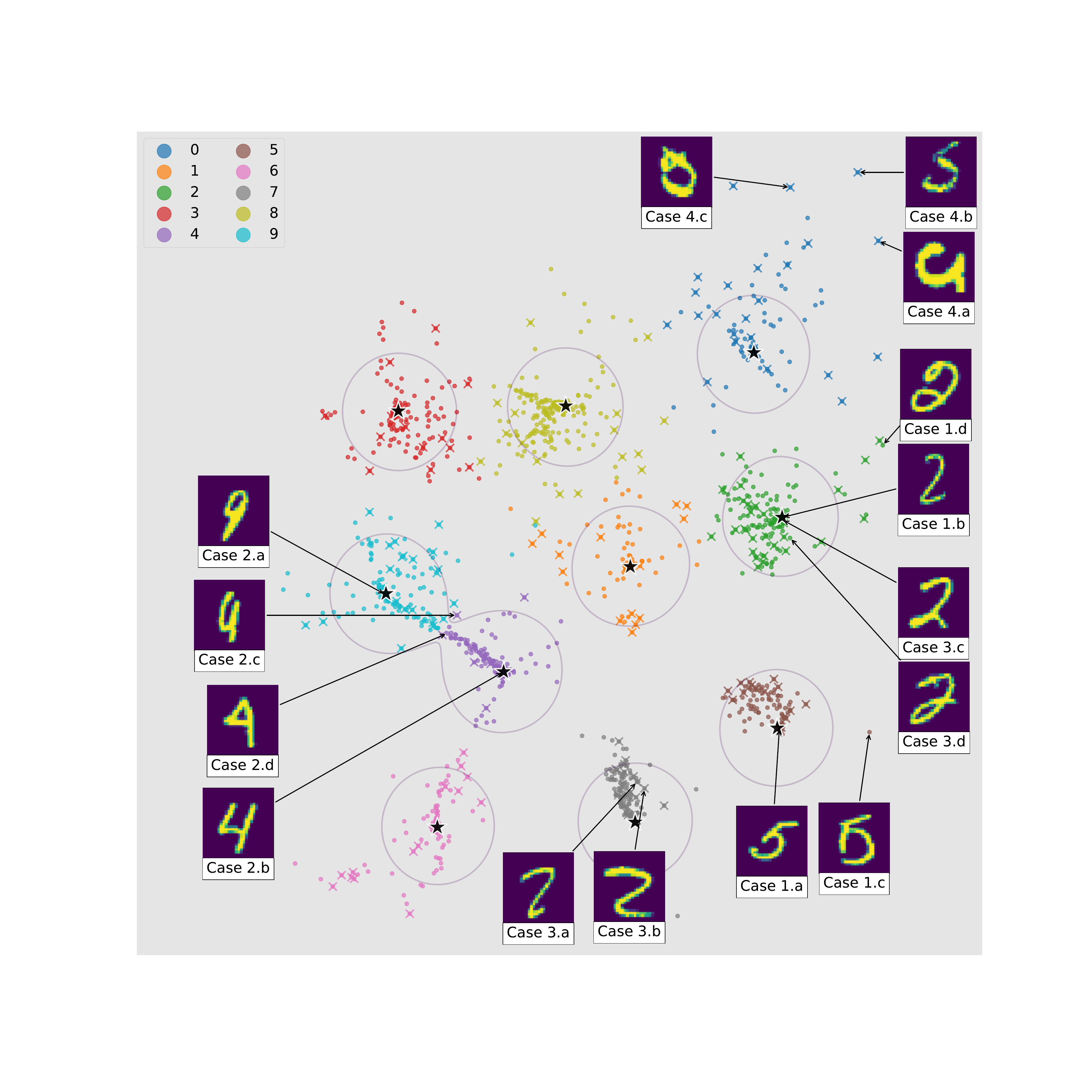} \label{fig:res-mnist}}}
            & \subfigure[VGG16 on Cifar10]{
                    \includegraphics[clip, trim=1.7cm 1.7cm 1.7cm 1.7cm, width=.3\textwidth]{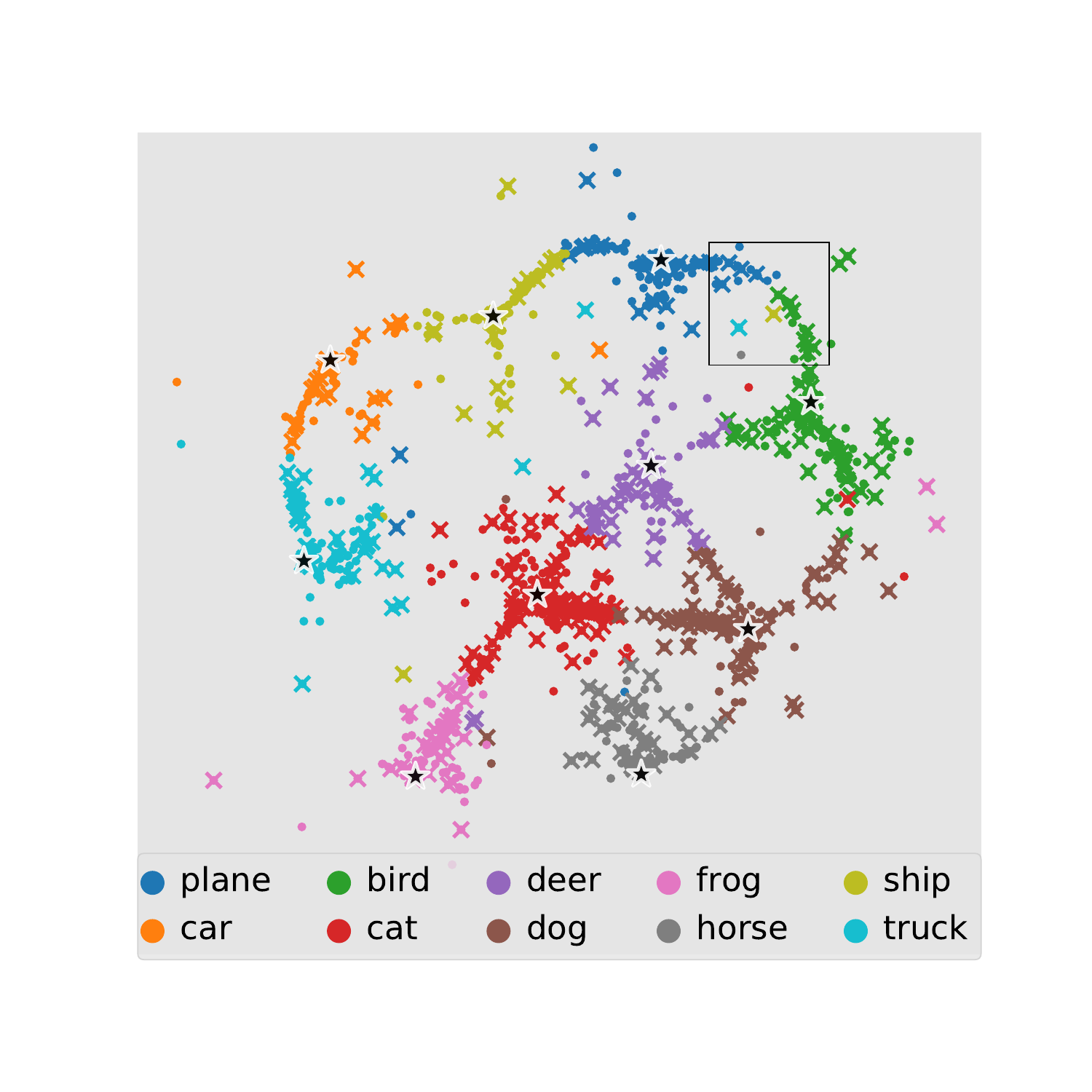} \label{fig:res-cifar10}
              } \\
            & \subfigure[Wide-ResNet on Cifar100]{
                \includegraphics[clip, trim=1.7cm 1.7cm 1.7cm 1.7cm, width=.3\textwidth]{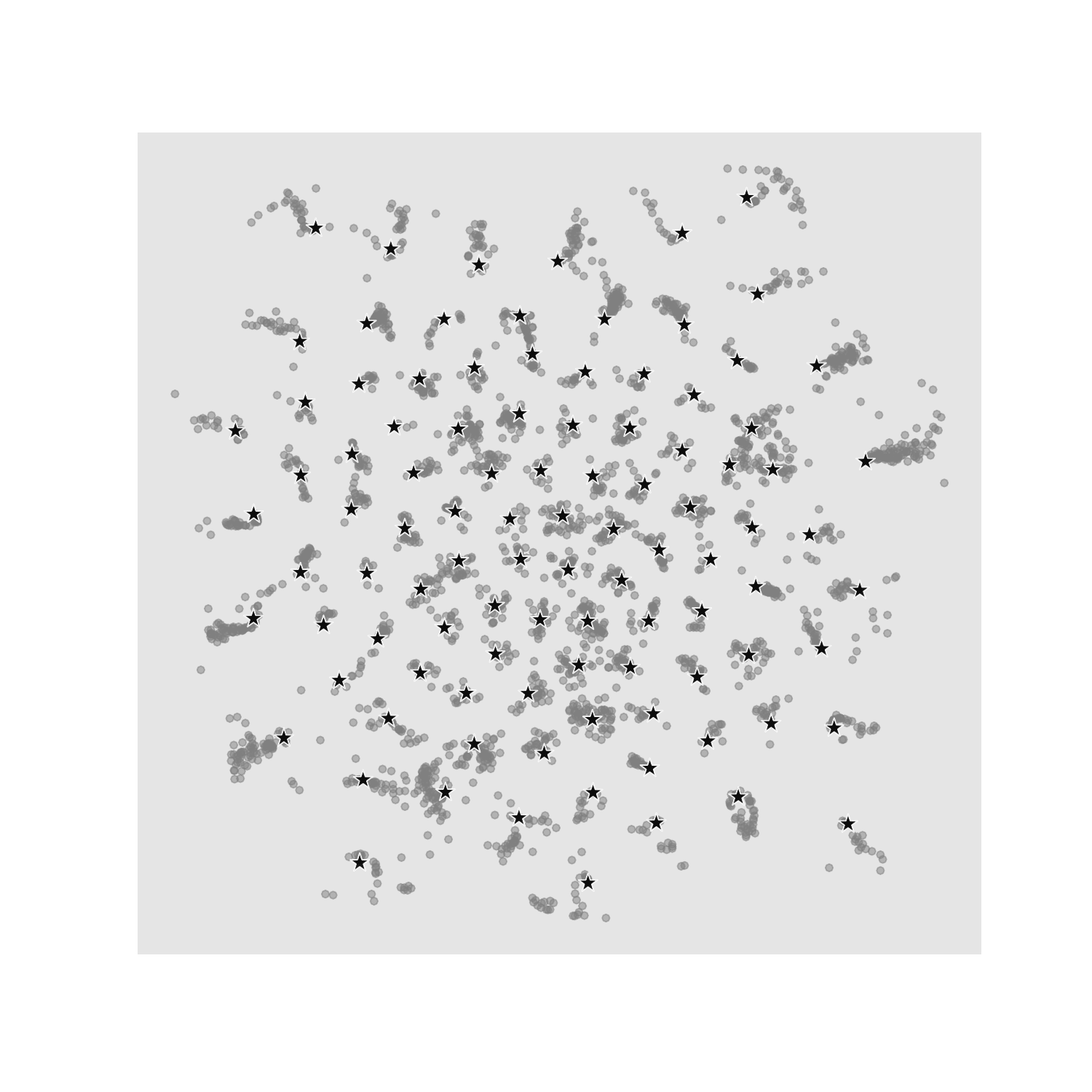}  \label{fig:res-cifar100}
              } \\
    \end{tabular}

    \caption
    {
        Scatter plots generated by DarkSight for LeNet (MNIST), VGG16 (Cifar10) and Wide-ResNet (Cifar100).
        For (a) and (b), points are colored by the teacher's predictions. 
        Rounded points means they are correctly classified by the teacher and crossings means they are wrongly classified by the teacher.
        For (c), we show the monochrome scatter plot simply because there are too many clusters which are hard to assign colors to.
        Stars in all plots are $\mu$s of the Student's t-distributions.
        In (a), the contour is where $P_S(y_i; \theta)$ equals to 0.001. 
    }
    \label{fig:results}
    \vskip -0.2in
\end{figure*}

In this section, we demonstrate the types of insights that a developer
of machine learning methods can gain
from examining DarkSight visualizations by giving example analyses on three data sets.
Figure~\ref{fig:results} shows the visualization generated by DarkSight for MNIST, Cifar10 and Cifar100.


\subsection{LeNet on MNIST}
\label{sec:res-mnist}
Figure~\ref{fig:res-mnist} visualizes the output of the MNIST classifier.
From this display, an analyst can gain several different types of insight into the classifier.
First, at a global level, we see that the points in the display are clustered into
points that have the same teacher prediction. Additionally, the locations of the clusters provide insight into which
classes seem most similar to the classifier, 
e.g. the clusters of points predicted as ``3'' and ``8'' are close together, ``4'' and ``9'' are close together, and so on.
Indeed, the relative distances between clusters can be interpreted by comparing
the contours of $P_S(y_i;\theta)$.
So for example, classes 
 ``4'' and ``9'' are the most similar pair, because
 their high-probability contours are the closest. By plotting more levels of contours, different levels of similarity between classes can be visualized.

Looking more deeply, the analyst can also see which
instances are most difficult for the classifier.
Because of DarkSight's cluster preservation property,
data points near a cluster centre have higher confidence, 
and points farther away have lower confidence.
This is illustrated by Case 1.
Here Case 1.a and 1.b indicate two points near the center of their clusters, which
 are typical-looking images.
In contrast, Case 1.c and 1.d indicate two digits which are far away from the centers
of the same cluster, and appear highly
 atypical.
The reason that DarkSight is able to display these
points as atypical is due to the dark knowledge inside 
their predictive probabilities.
For Case 1.c, the prediction vector is 
\texttt{["5":0.52 "0":0.44 \ldots]},
which contains an unusually high probability for the second-best class.
For Case 1.d, the prediction vector is 
\texttt{["2":0.52 "0":0.16 "9":0.12 "3":0.11 "8":0.08 \dots]}, which
contains an unusual number of classes with  large probability.
These two points are indicated as unusual in the visualization
because they lie at the edges of their clusters.



The cluster preservation property also implies 
that points midway between two clusters 
tend to be similar to both.
For example, Case 2.a is a typical ``9'' and Case 2.b is a typical ``4'',
and are located near the center of their respective clusters.
By contrast, Case 2.c and Case 2.d are midway between the two clusters, and both digits
are more difficult to recognize, with 
similarities to both ``9'' and ``4''.

Another interesting aspect of the visualization is that
nearby points that are misclassified tend to be misclassified in the same way.
For example, Case 3.a and Case 3.b, are ``2''s that are misclassified as 
``7''s.\footnote{
Instead of coloring points by their predictive labels,
one can also color them by their true labels, 
which makes it easy to spot misclassifications.
With the alternative coloring schema, an isolated red point, for example, in
a sea of blue points is likely to be a misclassification.
An example appears in the Appendix~\ref{app:color}.}
Compared with a typical ``2'' (e.g., Cases 1.d, 3.c and Case 3.d), 
these misclassified digits
seem to have a longer top horizontal stripe,
and lack the bottom curl.
This suggests what characteristics of digits 
are important for the classifier to make predictions.
One might consider improving the classifier 
for these inputs by either changing the architecture or collecting more examples
of this type of digit.

Finally, points appear as outliers 
 when the classifier predicts an
uncommon vector of predictive probabilities.
For example, Case 4 shows the three digits 
located in the upper-right corner of the plot.
It can be seen that these digits
are particularly unusual outliers. 
This suggests that a particularly interesting
class of anomalous data points are those
that cause a classifier to do unusual things.

\subsection{VGG16 on Cifar10}

Figure~\ref{fig:res-cifar10} shows the visualization for Cifar10.
In this plot, points are again grouped by top-predicted class, but now
six of the classes lie on a one-dimensional manifold that progresses
from ``truck'' to ``car'' through to ``bird'' and ``dog''. 
Along the curves connecting clusters, we observe that the top two probabilities in
the prediction vector smoothly transition through the classes in the manifold,
as we discussed in more detail earlier (see Figure~\ref{fig:black-box}).


\subsection{Wide-ResNet on Cifar100}

Finally, Figure~\ref{fig:res-cifar100} is the DarkSight visualization of the Wide-ResNet
trained on Cifar100 that includes all 100 classes. Even with so many classes,
it is possible to visually identify clusters and outliers.
Although it is difficult to examine a display with so many clusters in print,
an interactive plot that allows for panning and zoom can make 
it possible to explore this display thoroughly.

\section{Conclusions}


We present DarkSight,  a new dimension reduction technique for interpreting
deep classifiers based on 
knowledge distillation.
DarkSight jointly compresses a black-box classifier into a simpler, interpretable classifier and 
obtains the corresponding low-dimensional points for each input.
With DarkSight, one can faithfully visualise the predictive vectors 
from a classifier 
as shown through four useful properties.
We demonstrate how to use these properties to help diagnose deep classifiers, which
could potentially enable wider use of them in industry.


\ifdefined\isaccepted
\section*{Acknowledgements}

This work was funded by Edinburgh Huawei Research Lab, 
which is generously funded by Huawei Technologies Co. Ltd.
We thank Rich Caruana for his useful comments and suggestions on how to use DarkSight.
We also thank Cole Hurwitz for his careful grammar checking of the paper,
and Yiran Tao for his work on the demo of DarkSight. 
\fi




\bibliography{example_paper}
\bibliographystyle{icml2018}

\clearpage
\appendix

\section{Efficient Learning Algorithm for ``Simplified'' DarkSight by SVD}
\label{app:svd}

In a ``simplified'' version of DarkSight, 
i.e. DarkSight with softmax classifier as student model and 
MSE between logits as learning objective, 
there exists an efficient learning algorithm to generate 
the 2D embedding by SVD.

Firstly, we need to set up the notations
\begin{itemize}
    \item We call the output logits from teacher $L$, a N by K matrix where each row $L_i$ is the logit of input $i$;
    \item We call the low-dimensional points we would like to find $Y$, a N by 2 matrix where each row $Y_i$ represents the 2D location for input $i$;
    \item We parameterize the softmax classifier by a 2 by K matrix $W$
    \begin{itemize}
        \item The predictive probability is then $P_S(c_k|Y_i) = f_k(Y_i; W) = \text{Softmax}_k(Y_i W)$;
        \item The corresponding logits outputted by the student model is basically $S_i = Y_i W$;
    \end{itemize}
    \item The objective for all data points is
    \begin{equation}
        L(W, Y) = {||L - Y W||}_2^2.
        \label{eq:mse}      
    \end{equation}
\end{itemize}
The minimizing Equation~\ref{eq:mse} can also be formalized as: 
finding a decomposition $Y W$ for a N by K matrix $L$, where $Y$ is a N by 2 matrix and $W$ is a 2 by K matrix.
This is actually a low-rank matrix approximation problem, which can be found a solution by SVD.

In order to get the 2D embedding, we first perform a SVD on $L$ which gives
\begin{equation}
    L^T = U \Sigma V^T,
\end{equation}
where $U$ is a K by K matrix, $\Sigma$ is a K by K diagonal matrix and $V$ is a N by K matrix.
Then the best rank-2 approximation is simply $\tilde U \tilde \Sigma {\tilde V}^T$, where $\tilde U$ is the K by 2 submatrix of $U$, $\tilde \Sigma$ is $\Sigma$ with only the largest 2 singular values kept and $\tilde V$ is the N by 2 submatrix of $V$.
Therefore, the 2D embedding of DarkSight is $Y = \tilde V$ and the parameters of student model is $W = (\tilde U \tilde \Sigma)^T$.

\begin{figure}[ht]
\vskip 0.2in
\begin{center}
\centerline{\includegraphics[width=\columnwidth]{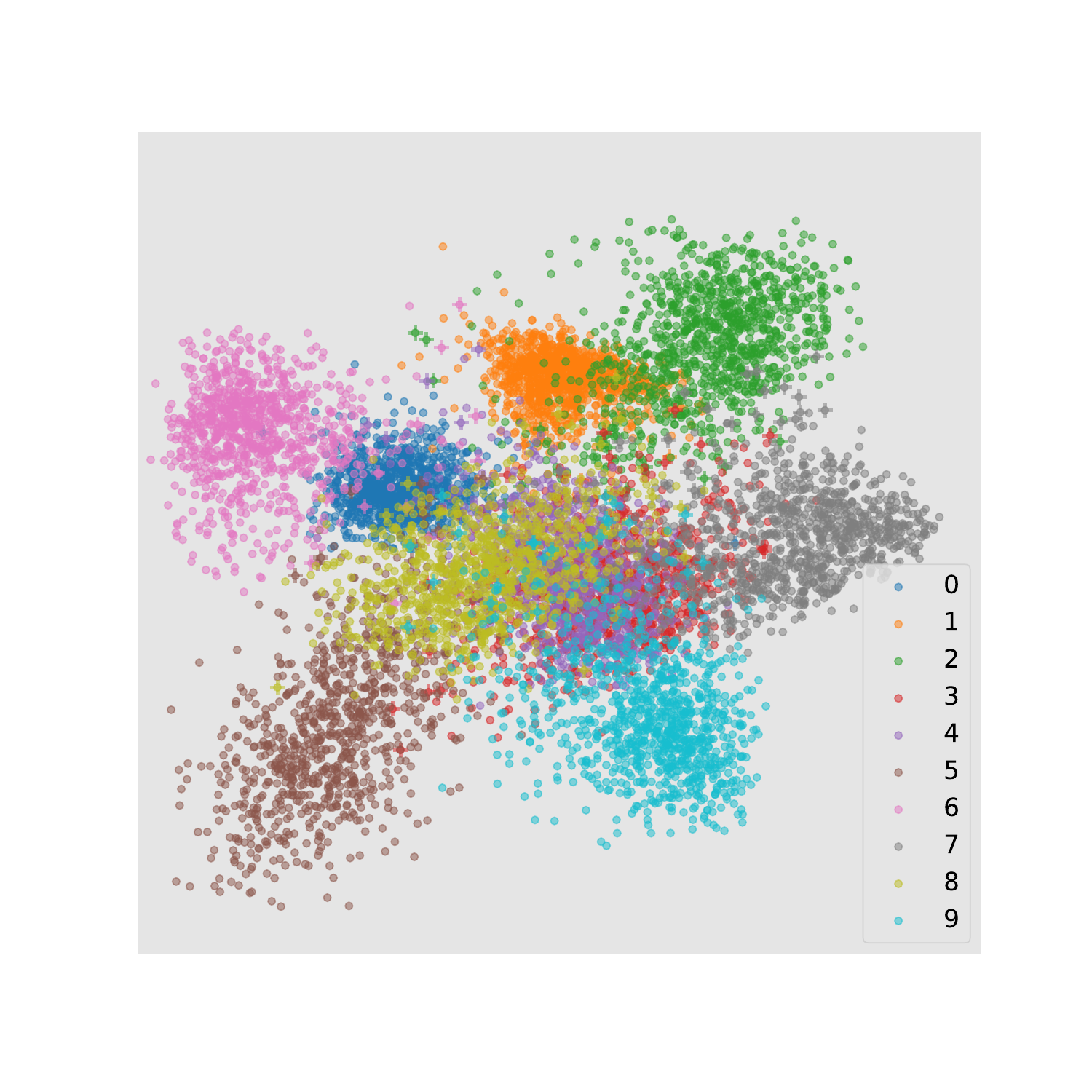}}
\caption{Visualization of LeNet on MNIST by ``Simplified'' DarkSight using SVD.}
\label{fig:mnist-svd}
\end{center}
\vskip -0.2in
\end{figure}

\begin{figure}[ht]
\vskip 0.2in
\begin{center}
\centerline{\includegraphics[width=\columnwidth]{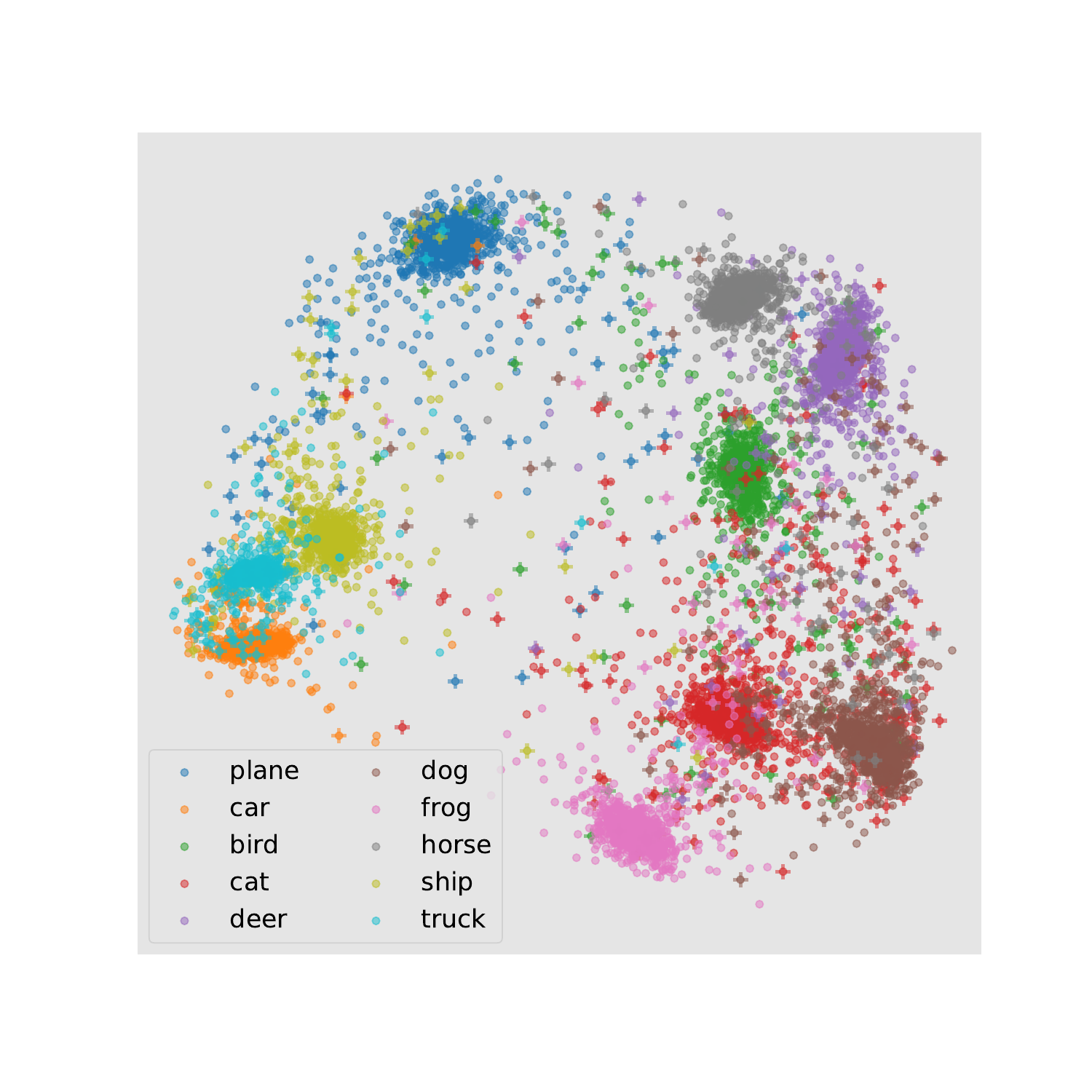}}
\caption{Visualization of VGG16 on Cifar10 by ``Simplified'' DarkSight using SVD.}
\label{fig:cifar10-svd}
\end{center}
\vskip -0.2in
\end{figure}

The visualization generated by the method described above of LeNet on MNIST and 
VGG16 on Cifar10 are shown in Figure~\ref{fig:mnist-svd} and Figure~\ref{fig:cifar10-svd} respectively.

\section{Variants of DarkSight}
\label{sec:app-variants}
Besides the basic setting of DarkSight introduce in Section~\ref{sec:darksight}, 
there are three useful variants of DarkSight for particular purposes.

\subsection{Parametric DarkSight}
\label{sec:app-variants-param}

The DarkSight method presented in Section~\ref{sec:darksight} is non-parametric
because the parameters
include the low-dimensional embedding \protect $y_i$ for each data point \protect $x_i$, 
and hence grow linearly with the amount of data. 
In some situations, it might be more desirable to have a parametric version.
Motivated by \citet{maaten2009learning}, we suggest one can train a map $f_p$ from $P_T(\cdot|x_i)$ to $y_i$ using a neural network.
It could be done by firstly training a non-parametric DarkSight to obtain $Y$ and 
then fitting $f_p$ to minimize $L(p) = \frac{1}{N} \sum_{i=1}^{N} (f(P_T(\cdot|x_i)) - y_i)^2$, followed by an end-to-end fine-tuning\footnote{The end-to-end fine-tuning can by done by replacing each $y_i$ by $f_p(P_T(\cdot|x_i))$ in Equation~\ref{eq:obj}, thus the parameters to optimize are $\theta$ and $p$. The training algorithm is similar: either plain SGD or coordinate descent by SGD can be used to optmize $\theta$ and $p$.}.

This actually enables an interesting pipeline to generate heatmap visualization.
Recall that from the student model we have $P(y_i;\theta)$.
The term is small if $y_i$ is far away from clusters, 
so it can be used as an alternative confidence measure rather than the one based on density estimation.
With the parametric DarkSight, one can get this value by $P(f_p(P_T(\cdot|x_i);\theta)$,
which is differentiable w.r.t the input $x_i$.
Thus, one can generate a heatmap based on the gradient of $P(y_i;\theta)$ w.r.t $x_i$, i.e. $\nabla_{x_i} P(f_p(P_T(\cdot|x_i);\theta)$.
A further interesting work is to generate adversarial inputs based on this gradient.

\subsection{DarkSight for Subset of Classes}
\label{sec:app-variants-sub}
Although it is technically possible to visualize classifiers with 100 classes by DarkSight 
(
see Table~\ref{tab:results} and Figure~\ref{fig:res-cifar100} 
), 
two clusters of interest might be placed far away from each other
, in which case other clusters
in between would prevent interesting visual interactions between clusters.
To tackle this problem, one can choose to visualize only a subset of interesting classes.

In order to do that, the \textit{normalized subset of predictive distribution} can be used as learning target 
\begin{equation}
    P^{\text{Sub}}_t(\cdot|x_i) = \frac{P_T(\cdot|x_i) \odot \mathbf{m}}{\sum_{i=1}^{N} (P_T(\cdot|x_i) \odot \mathbf{m})},
\end{equation}
where $\odot$ is the element-wise product and $\mathbf{m}$ is a length-$K$ vector with 1 indicating classes to visualize and 0 not to.

Note that with this variant, one can choose to visualize the whole dataset by creating a small multiple, 
i.e. grid plot of 100 classes by 10 plots with 10 classes each.

\subsection{Control the size of high confidence area}
\label{sec:app-variants-size}
In DarkSight, points with high confidence are located in a small region while points with low confidence spread around,
which is an interesting phenomena of our method which allows outliers to be detected.
On other side, high confident points are usually more than those with low confidence, 
which makes most of points located in a small area.
Thus, one might be interested in control the degree of such phenomena.
Inspired by \citep{hinton2015distilling}, it can be achieved by introducing a temperature term $T$ 
when normalizing logits $l$ to probability $p$:
    $p = l' / \sum_i l'$, where $l' = l_i / T$.
By setting $T$ greater than 1, one can encourage the points in the 
visualizatoin to spread more apart, which further pulls put outliers from clusters.

Setting $T$ greater than 1 also makes the knowledge distillation 
easier as it increases the entropy of $p$, i.e. the information provided by the teacher.
Therefore when training DarkSight, annealing can be used by firstly set $T$ to be a high value, e.g. 20,
and generally reducing it to 1.

\section{Experimental Setups}
\label{app:exp-setup}
In all experiments, optimization is run for 1,000 epochs by an Adam optimizer with a batch size of 1,000.
MNIST and Cifar100 uses coordinate descent by SGD and Cifar10 uses plain SGD with annealing based on temperature mentioned in Appendix~\ref{sec:app-variants-size}.

The covariance matrices of the conditional distribution are initialized as 
$\sqrt{\log K} I$ and are 
not optimized.
We found that DarkSight is able to learn a fairly good mimic even without optimizing the covariance. Also, fixing the covariance (or relaxing it to be $\lambda I$) makes the plot easy to interpret.
The means of the conditional distributions are randomly initialized by 2D Gaussian.
For Cifar10, they are randomly initialized by 2D Gaussian; for MNIST and Cifar100, all low-dimensional points are initialized at the centers of their predictive cluster.

For the Student's t distribution, we use 
$\nu = 2$  degrees of freedom.
We use different learning rates for different parameters. 
For MNIST and Cifar10, we use the learning rate $\eta_{c} = 0.001$ for $\theta_c$; $\eta_{p} = 0.005$ for $\theta_p$; and $\eta_{Y} = 1 \times 10^{-6}$
for $y$. For Cifar100,  we use $\eta_{c} = 0.005$, $\eta_{p} = 0.01$ and $\eta_{Y} = 1 \times 10^{-6}$.

\section{Running Time Benchmarks}
\label{app:time}
DarkSight visualization for a dataset with 10,000 data points and 10 classes 
can be generated by coordinate descent in 12 minutes with CPU only and 1.5 minutes with GPU;
or by plain SGD in 6 minutes with CPU only and 48 seconds with GPU.
For t-SNE on the same dataset and dimensionality, it costs around 12 minutes with CPU only.
The t-SNE we use is a Julia implementation of t-SNE, available at \url{https://github.com/lejon/TSne.jl}.
Julia is believed to be more efficient than MATLAB or Python in terms of scientific computing on CPUs.

Recall that as DarkSight has a time complexity of $O(N)$. 
In order to visualize more than 10,000 points,
t-SNE, with a time complexity of $O(N^2)$ or $O(N\log N)$, would spend more time than DarkSight.

\section{Alternative Color Schema}
\label{app:color}
\begin{figure*}[ht]
    \vskip 0.2in
    \begin{center}
        \includegraphics[width=.69\textwidth]{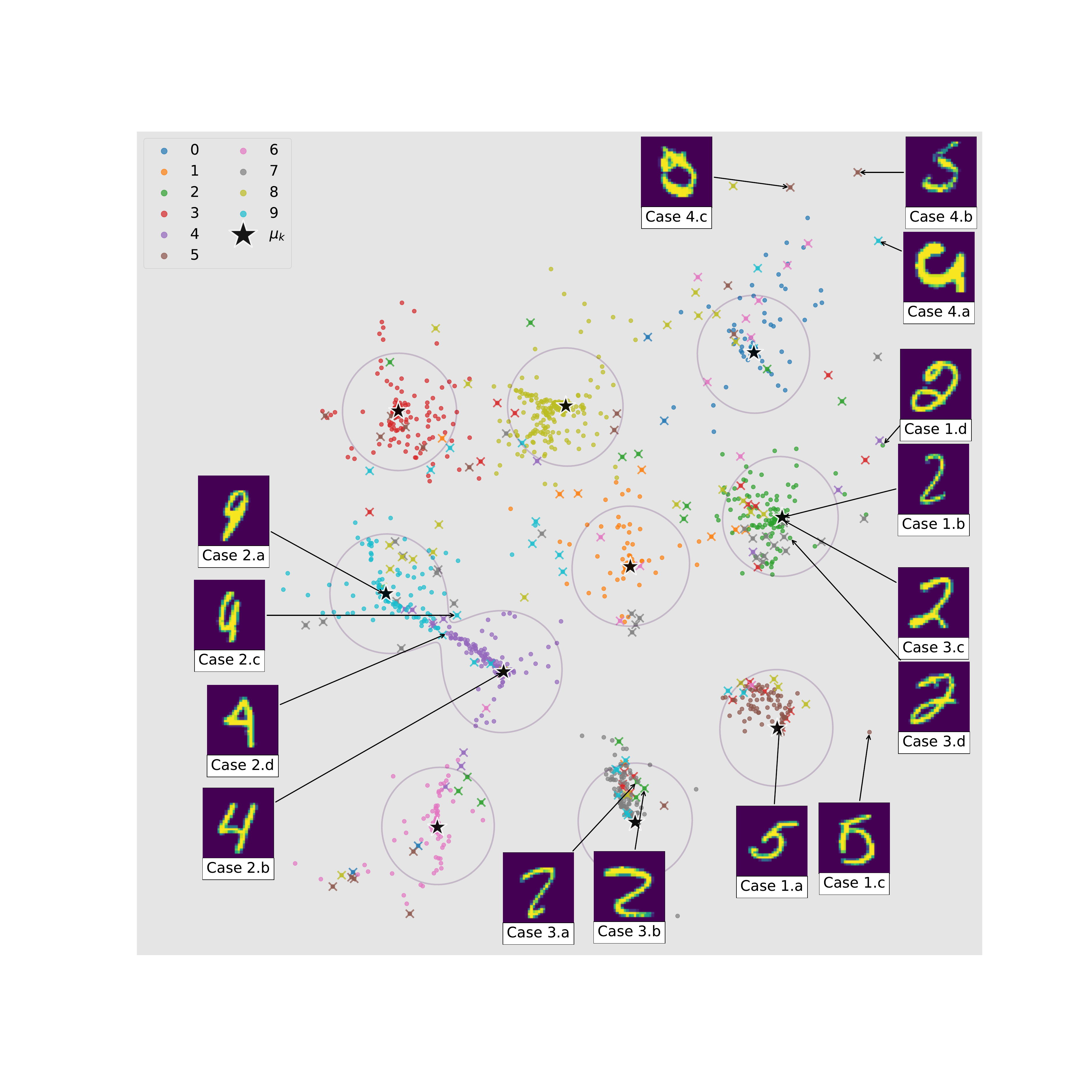}
        \end{center}
    \vskip -0.2in
    \caption{Same visualization as Figure~\ref{fig:res-mnist} with points colored by their true labels.} 
    \label{fig:res-mnist-color}
\end{figure*}

Except from coloring points by their predictive labels, one can also color points by their true labels.
Figure~\ref{fig:res-mnist-color} shows the same low-dimensional embedding as Figure~\ref{fig:res-mnist} with this alternative color schema.

\section{Confusion Matrix for VGG16 on Cifar10}
\label{sec:conf-mat-cifar10}
Table~\ref{tab:conf-mat-cifar10} shows the confusion matrix for VGG16 on Cifar10.
\begin{table*}[ht]
\centering
\caption{Confusion Matrix for VGG16 on Cifar10. Rows are predictive labels and columns are true labels.}
\label{tab:conf-mat-cifar10}
\begin{tabular}{rcccccccccc}
 & plane                    & car                      & bird                     & cat                      & deer                     & dog                      & frog                     & horse                    & ship                     & truck                    \\ \cline{2-11} 
\multicolumn{1}{r|}{plane}      & \multicolumn{1}{c|}{951} & \multicolumn{1}{c|}{2}   & \multicolumn{1}{c|}{11}  & \multicolumn{1}{c|}{6}   & \multicolumn{1}{c|}{1}   & \multicolumn{1}{c|}{2}   & \multicolumn{1}{c|}{4}   & \multicolumn{1}{c|}{2}   & \multicolumn{1}{c|}{16}  & \multicolumn{1}{c|}{5}   \\ \cline{2-11} 
\multicolumn{1}{r|}{car}        & \multicolumn{1}{c|}{3}   & \multicolumn{1}{c|}{972} & \multicolumn{1}{c|}{0}   & \multicolumn{1}{c|}{1}   & \multicolumn{1}{c|}{0}   & \multicolumn{1}{c|}{1}   & \multicolumn{1}{c|}{0}   & \multicolumn{1}{c|}{0}   & \multicolumn{1}{c|}{5}   & \multicolumn{1}{c|}{18}  \\ \cline{2-11} 
\multicolumn{1}{r|}{bird}       & \multicolumn{1}{c|}{13}  & \multicolumn{1}{c|}{0}   & \multicolumn{1}{c|}{920} & \multicolumn{1}{c|}{16}  & \multicolumn{1}{c|}{10}  & \multicolumn{1}{c|}{12}  & \multicolumn{1}{c|}{11}  & \multicolumn{1}{c|}{5}   & \multicolumn{1}{c|}{2}   & \multicolumn{1}{c|}{1}   \\ \cline{2-11} 
\multicolumn{1}{r|}{cat}        & \multicolumn{1}{c|}{5}   & \multicolumn{1}{c|}{0}   & \multicolumn{1}{c|}{17}  & \multicolumn{1}{c|}{852} & \multicolumn{1}{c|}{12}  & \multicolumn{1}{c|}{57}  & \multicolumn{1}{c|}{12}  & \multicolumn{1}{c|}{8}   & \multicolumn{1}{c|}{3}   & \multicolumn{1}{c|}{2}   \\ \cline{2-11} 
\multicolumn{1}{r|}{deer}       & \multicolumn{1}{c|}{4}   & \multicolumn{1}{c|}{0}   & \multicolumn{1}{c|}{14}  & \multicolumn{1}{c|}{19}  & \multicolumn{1}{c|}{956} & \multicolumn{1}{c|}{15}  & \multicolumn{1}{c|}{8}   & \multicolumn{1}{c|}{10}  & \multicolumn{1}{c|}{0}   & \multicolumn{1}{c|}{0}   \\ \cline{2-11} 
\multicolumn{1}{r|}{dog}        & \multicolumn{1}{c|}{0}   & \multicolumn{1}{c|}{0}   & \multicolumn{1}{c|}{16}  & \multicolumn{1}{c|}{76}  & \multicolumn{1}{c|}{7}   & \multicolumn{1}{c|}{894} & \multicolumn{1}{c|}{2}   & \multicolumn{1}{c|}{11}  & \multicolumn{1}{c|}{0}   & \multicolumn{1}{c|}{1}   \\ \cline{2-11} 
\multicolumn{1}{r|}{frog}       & \multicolumn{1}{c|}{2}   & \multicolumn{1}{c|}{1}   & \multicolumn{1}{c|}{11}  & \multicolumn{1}{c|}{19}  & \multicolumn{1}{c|}{4}   & \multicolumn{1}{c|}{3}   & \multicolumn{1}{c|}{961} & \multicolumn{1}{c|}{0}   & \multicolumn{1}{c|}{2}   & \multicolumn{1}{c|}{0}   \\ \cline{2-11} 
\multicolumn{1}{r|}{horse}      & \multicolumn{1}{c|}{1}   & \multicolumn{1}{c|}{0}   & \multicolumn{1}{c|}{5}   & \multicolumn{1}{c|}{3}   & \multicolumn{1}{c|}{10}  & \multicolumn{1}{c|}{14}  & \multicolumn{1}{c|}{1}   & \multicolumn{1}{c|}{962} & \multicolumn{1}{c|}{0}   & \multicolumn{1}{c|}{0}   \\ \cline{2-11} 
\multicolumn{1}{r|}{ship}       & \multicolumn{1}{c|}{14}  & \multicolumn{1}{c|}{3}   & \multicolumn{1}{c|}{4}   & \multicolumn{1}{c|}{4}   & \multicolumn{1}{c|}{0}   & \multicolumn{1}{c|}{0}   & \multicolumn{1}{c|}{0}   & \multicolumn{1}{c|}{0}   & \multicolumn{1}{c|}{964} & \multicolumn{1}{c|}{4}   \\ \cline{2-11} 
\multicolumn{1}{r|}{truck}      & \multicolumn{1}{c|}{7}   & \multicolumn{1}{c|}{22}  & \multicolumn{1}{c|}{2}   & \multicolumn{1}{c|}{4}   & \multicolumn{1}{c|}{0}   & \multicolumn{1}{c|}{2}   & \multicolumn{1}{c|}{1}   & \multicolumn{1}{c|}{2}   & \multicolumn{1}{c|}{8}   & \multicolumn{1}{c|}{969} \\ \cline{2-11} 
\end{tabular}
\end{table*}

\end{document}